\documentclass[10pt,a4paper,twoside,twocolumn]{article}
\pdfoutput=1

\usepackage{cmpproc}

\usepackage{amsfonts}
\usepackage[fleqn]{amsmath}
\usepackage{amssymb}
\usepackage{amsthm}

\newcommand{\sm}[1]{\ensuremath{\mathtt{\uppercase{#1}}}}
\newcommand{\sv}[1]{\ensuremath{\mathbf{\lowercase{#1}}}}
\newcommand{\SO}[1]{\operatorname{SO}\left(#1\right)}
\newcommand{\norm}[1]{\ensuremath{\left\Vert #1 \right\Vert}}
\newcommand{\transpose}{\ensuremath{^\mathsf{T}}}
\newcommand{\coloneqq}{\mathrel{\mathop:}=}
\newcommand{\ah}[0]{{\alpha^\prime}}
\newcommand{\bh}[0]{{\beta^\prime}}
\newcommand{\rod}{\mathcal{R}}

\newtheoremstyle{dotless}{}{}{\itshape}{}{\bfseries}{}{ }{}
\theoremstyle{dotless}
\newtheorem{lemma}{Lemma}

\newtheorem{definition}{Definition}

\usepackage[backend=bibtex]{biblatex}
\usepackage{graphicx}
\usepackage{hyperref}
\usepackage{balance}

\begin{document}

\title{On the Relation between two Rotation Metrics}
\subtitle{}
\author{Ruland, Thomas}
\affiliation{%
   Ulm University\\
   \href{mailto:thomas.ruland@uni-ulm.de}{thomas.ruland@uni-ulm.de}
}

\begin{proceeding}

\begin{abstract}
In their work "Global Optimization through Rotation Space Search" \cite{Hartley2009}, Richard Hartley and Fredrik Kahl introduce a global optimization strategy for problems in geometric computer vision, based on rotation space search using a branch-and-bound algorithm. In its core, Lemma~2 of their publication is the important foundation for a class of global optimization algorithms, which is adopted over a wide range of problems in subsequent publications. This lemma relates a metric on rotations represented by rotation matrices with a metric on rotations in axis-angle representation. This work focuses on a proof for this relationship, which is based on Rodrigues' Rotation Theorem for the composition of rotations in axis-angle representation \cite{rodrigues1840lois,altmann1989hamilton}.
\end{abstract}

\section{Introduction}

In geometry, various representations exist to describe a rotation in Euclidean 3-space. The focus of this work is the relationship between two metrics on the following two rotation representations.\begin{description}
\item[Rotation Matrices] The group of isometric, linear transformations which preserve handedness in space is referred to as $\SO{3} \coloneqq \left\{ \sm{R} \in \mathbb{R}^{3\times3} \vert \sm{R}\transpose\sm{R} = \sm{I}, \ \det\sm{R} = 1 \right\}$.
\item[Multiplied Axis-Angle] $\rod \subset \mathbb{R}^3$ denotes the space of multiplied axis-angle\footnote{Also know as Rodrigues parameters, named after  Benjamin Olinde Rodrigues (1795-1851)} representations $\sv{r} = \alpha \ \sv{a}$ of rotations with angle $\alpha \in \left[0, \pi\right]$ about the axis $\sv{a} \in \mathbb{R}^3$ with $\norm{\sv{a}} = 1$. It describes a closed ball of radius $\pi$ in $\mathbb{R}^3$.
\end{description}

Geometric computer vision or reconstruction problems are often formulated as a task of minimizing a cost or objective function. In general, these objective functions are non-convex. As a result, standard local optimization algorithms only yield locally optimal results. Hartley and Kahl contributed a global optimization strategy for problems in geometric computer vision, based on rotation space search using a branch-and-bound algorithm \cite{Hartley2009}. Their work is the foundation for several subsequent contributions \cite{bazin2013globally,choi2010branch,HartleyRotAve,Heller2012a,7206586,Ruland2012CVPR,Seo2009}. Hartley and Kahl's key contribution, which enables the branch-and-bound search strategy, is Lemma~\ref{lemma::relationSO3Rodrigues}. It relates a metric on rotations represented by rotation matrices from $\SO{3}$ with a metric on rotations in multiplied axis-angle representation $\rod$.

\section{Previous Work}

This section restates definitions and lemmas from \cite{Hartley2009}. Definition~\ref{def::metricSO3} introduces a metric on the space $\SO{3}$ of rotation matrices.

\begin{definition}
\label{def::metricSO3}
Let the rotation matrices $\sm{R}_A, \sm{R}_B \in \SO{3}$ represent two rotations in Euclidean 3-space. The operation $\operatorname{d}_\angle: \SO{3} \times \SO{3} \to \left[0, \pi\right]$ defines the metric\begin{align}
\operatorname{d}_\angle\left(\sm{R}_A, \sm{R}_B\right) \coloneqq \angle\left(\sm{R}_B^{-1} \sm{R}_A\right),
\end{align}where the angle operator $\angle\left(\cdot\right)$ yields the rotation angle of the given rotation matrix after decomposition into rotation axis and angle.
\end{definition}

Lemma~\ref{lemma::relationSO3Rodrigues} relates this metric to the Euclidean distance of the respective rotations in multiplied axis-angle representation.

\begin{lemma}
\label{lemma::relationSO3Rodrigues}
Let two rotations about axes $\sv{a}_A, \sv{a}_B \in \mathbb{R}^3$, $\norm{\sv{a}_A} = \norm{\sv{a}_B} = 1$ by the angles $\alpha, \beta \in \left[0, \pi\right]$ be represented by the multiplied axis-angle representations $\sv{r}_A = \alpha \ \sv{a}_A \in \mathcal{R}$ and $\sv{r}_B = \beta \ \sv{a}_B \in \mathcal{R}$ as well as the rotation matrices $\sm{R}_A, \sm{R}_B \in \SO{3}$. The following relationship holds:\begin{align}
\label{eqn::theUnequality}
\operatorname{d}_\angle\left(\sm{R}_A, \sm{R}_B\right) \leq \norm{\sv{r}_A - \sv{r}_B}.
\end{align}
\end{lemma}

This work focuses on a proof for Lemma~\ref{lemma::relationSO3Rodrigues}.

\section{The Proof}

The key idea of this proof is to cast Lemma~\ref{lemma::relationSO3Rodrigues} to an upper bound on the rotation angle of the composed rotation $\sm{R}_B \sm{R}_A$. To achieve this, all representations of rotation $B$ are inverted without loss of generality. $B$'s multiplied axis-angle representation $\sv{r}_B$ is substituted by $-\sv{r}_B$ and its rotation matrix representation $\sm{R}_B$ by $\sm{R}_B^{-1}$, respectively:\begin{align}
\label{eqn::rotationCompositionInequ}
\operatorname{d}_\angle\left(\sm{R}_A, \sm{R}_B^{-1}\right) \leq \norm{\sv{r}_A - \left(-\sv{r}_B\right)}.
\end{align}The verification of this upper bound is grouped into three sections. Section~\ref{sec::part1} reformulates the left hand side fully in terms of the three angles $\alpha, \beta$ and $\varphi$, where $\varphi$ denotes the angle enclosed by $\sv{a}_A$ and $\sv{a}_B$\begin{align}
\varphi = \arccos \sv{a}_A\transpose \sv{a}_B.
\end{align}Section~\ref{sec::part2} follows the same goal for the right hand side. Section~\ref{sec::part3} then verifies inequality \eqref{eqn::rotationCompositionInequ} by reducing it to a test for function convexity.

\subsection{The Left Hand Side}
\label{sec::part1}

\graphicspath{{images/}}
\begin{figure}
        \centering
                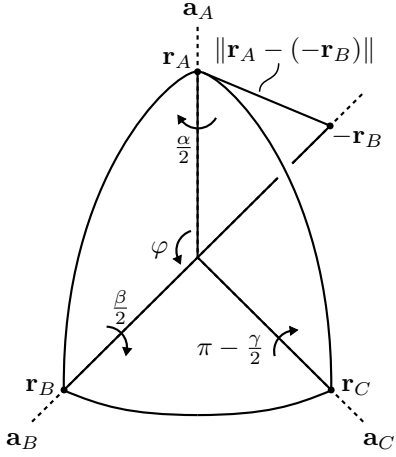
                \caption{Geometrical illustration of Rodrigues' rotation theorem on the composition of rotations in axis-angle representation \cite{altmann1989hamilton,rodrigues1840lois}. The theorem studies the spherical triangle spanned by the rotation axes $\sv{a}_A, \sv{a}_B$ and $\sv{a}_C$. In this triangle, $\sv{a}_C$ and $\gamma$ are the axis and the angle of the composed rotation $B \circ A$. The three corresponding rotation angles $\alpha, \beta$ and $\gamma$ appear as half-angles.}
                \label{img::rodRotTheo}
\end{figure}

Applying Definition~\ref{def::metricSO3} to the rotation matrix metric $\operatorname{d}_\angle$ on the left hand side of \eqref{eqn::rotationCompositionInequ} yields\begin{align}
\operatorname{d}_\angle\left(\sm{R}_A, \sm{R}_B^{-1}\right) = \angle\left(\sm{R}_B \sm{R}_A\right),
\end{align}the angle of the composed rotation $B \circ A$. The goal of this section is to express this composition directly in $\rod$. This is enabled by Rodrigues' Rotation Theorem on the composition of rotations in axis-angle representation \cite{altmann1989hamilton,rodrigues1840lois}. Figure~\ref{img::rodRotTheo} illustrates its geometrical interpretation. The theorem provides a closed form solution for the angle\begin{align}
& \angle\left(\sm{R}_B \sm{R}_A\right) \nonumber
\\
\label{eqn::rodriguesRotationTheorem}
& = 2\arccos\left( \cos\frac{\alpha}{2} \cos\frac{\beta}{2}-\sin\frac{\alpha}{2} \sin\frac{\beta}{2} \cos\varphi \right).
\end{align}By substituting $1-2d = \cos\varphi$, $\ah = \frac{\alpha}{2}$ and $\bh = \frac{\beta}{2}$, the argument of the $\arccos$ in \eqref{eqn::rodriguesRotationTheorem} expands to\begin{align}
& \cos\ah\cos\bh - \left(1-2d\right) \sin\ah\sin\bh
\\
& = \cos\ah\cos\bh - \sin\ah\sin\bh 
\\ & \quad + d \sin\ah\sin\bh + d \sin\ah\sin\bh \nonumber
\\
& = \cos\ah\cos\bh - \sin\ah\sin\bh 
\\ & \quad + d \sin\ah\sin\bh + d \sin\ah\sin\bh \nonumber
\\ & \quad + d \cos\ah\cos\bh - d \cos\ah\cos\bh. \nonumber
\end{align}Applying the trigonometric addition and subtraction theorem $\cos\left(\alpha\pm\beta\right) = \cos\alpha \cos\beta \mp \sin\alpha \sin\beta$ simplifies this argument to\begin{align}
& d \cos\left(\ah - \bh\right) - d \cos\left(\ah + \bh\right) + \cos\left(\ah + \bh\right) \nonumber
\\
& = d \cos\left(\ah - \bh\right) + \left(1-d\right) \cos\left(\ah + \bh\right).
\end{align}In summary, for the left hand side of \eqref{eqn::rotationCompositionInequ} holds\begin{align}
& \angle\left(\sm{R}_B \sm{R}_A\right)
\\
& = 2\arccos\left(d \cos\left(\ah - \bh\right) + \left(1-d\right) \cos\left(\ah + \bh\right)\right). \nonumber
\end{align}

\subsection{The Right Hand Side}
\label{sec::part2}

Figure~\ref{img::rodRotTheo} illustrates the expression $\norm{\sv{r}_A - \left(-\sv{r}_B\right)}$, on the right hand side of \eqref{eqn::rotationCompositionInequ}. When constructing the triangle spanned by the origin $\sv{0}$, $\sv{r}_A$ and $-\sv{r}_B$, the length of two sides ($\norm{\sv{r}_A}$ and ($\norm{-\sv{r}_B}$) and the angle enclosed by these two sides ($\pi - \varphi$) is known. The law of cosines yields the length of the remaining side\begin{align}
& \norm{\sv{r}_A - \left(-\sv{r}_B\right)} \nonumber
\\
& = \sqrt{\norm{\sv{r}_A}^2 - 2 \norm{\sv{r}_A} \norm{\sv{r}_B} \cos\left(\pi - \varphi\right) + \norm{\sv{r}_B}^2}.
\end{align}By definition of $\sv{r}_A$ and $\sv{r}_B$ and the symmetry of the cosine, this is equal to\begin{align}
2 \sqrt{\left(\frac{\alpha}{2}\right)^2 - 2 \frac{\alpha}{2} \frac{\beta}{2} \left(-\cos \varphi\right) + \left(\frac{\beta}{2}\right)^2}.
\end{align}Again substituting $1-2d = \cos\varphi$, $\ah = \frac{\alpha}{2}$ and $\bh = \frac{\beta}{2}$ yields for the argument of the square root\begin{align}
& \ah^2 - 2 \ah \bh \left(2d-1\right) + \bh^2
\\
& = \ah^2 - 2 d \ah\bh - 2 d \ah\bh + 2 \ah\bh + \bh^2
\\
& = \ah^2 - 2 d \ah\bh - 2 d \ah\bh + 2 \ah\bh + \bh^2 \nonumber
\\
& \quad + d \ah^2 - d \ah^2 + d \bh^2 - d \bh^2
\\
& = d \left(\ah^2 - 2\ah\bh + \bh^2\right) \nonumber
\\
& \quad + \left(\ah^2 + 2\ah\bh + \bh^2\right) \nonumber
\\
& \quad - d \left(\ah^2 + 2\ah\bh + \bh^2\right)
\\
& = d \left(\ah - \bh\right)^2 + \left(1-d\right) \left(\ah + \bh\right)^2.
\end{align}In total, for the right hand side of \eqref{eqn::rotationCompositionInequ} it holds\begin{align}
& \norm{\sv{r}_A - \left(-\sv{r}_B\right)} \nonumber
\\
& = 2 \sqrt{d \left(\ah - \bh\right)^2 + \left(1-d\right) \left(\ah + \bh\right)^2}
\end{align}

\subsection{Verifying the Inequality}
\label{sec::part3}

Collecting both sides, \eqref{eqn::rotationCompositionInequ} now is transformed to\begin{align}
& \arccos\left(d \cos\left(\ah - \bh\right) + \left(1-d\right) \cos\left(\ah + \bh\right)\right)
\nonumber
\\
& \ \ \leq \sqrt{d \left(\ah - \bh\right)^2 + \left(1-d\right) \left(\ah + \bh\right)^2}.
\end{align}This relationship is strongly related to the property of relative convexity \cite{Palmer2003} and in the following is examined in a similar way. Since both $\arccos$ and the square root are non-negative functions, the inequality still holds for the square of both sides\begin{align}
& \arccos^2\left(d \cos a + \left(1-d\right) \cos b\right)
\nonumber
\\
& \ \leq d \ a^2 + \left(1-d\right) b^2,
\end{align}where $a = \ah - \bh$ and $b = \ah + \bh$. Both $\ah$ and $\bh$ are limited to the interval $\left[0, \frac{\pi}{2}\right]$. The derived parameters $a$ and $b$ are thus within $\left[-\frac{\pi}{2}, \frac{\pi}{2}\right]$. On this interval, it is safe to substitute $a^2 = \left(\arccos \cos a\right)^2$ and $b^2 = \left(\arccos \cos b\right)^2$ to result in\begin{align}
& \arccos^2\left(d \cos a + \left(1-d\right) \cos b\right)
\nonumber
\\
& \ \leq d \left(\arccos \cos a\right)^2 + \left(1-d\right) \left(\arccos \cos b\right)^2.
\end{align}Finally, simplifying the inequality by substituting $a^\prime = \cos a$ and $b^\prime = \cos b$ yields\begin{align}
& \arccos^2\left(d \ a^\prime + \left(1-d\right) b^\prime\right)
\nonumber
\\
& \ \leq d \arccos^2 a^\prime + \left(1-d\right) \arccos^2 b^\prime.
\end{align}This reduces the proof to verifying the convexity of the function $\arccos^2$ on the interval $\left[0, 1\right]$. For $\operatorname{f}\left(x\right) = \arccos^2 x$ to be convex, its second derivative\begin{align}
\operatorname{f}^{\prime\prime}\left(x\right) = \frac{2 \sqrt{1 - x^2} - 2 x \arccos x}
{(1 - x^2)^{3/2}}
\end{align}has to be non-negative. The denominator of $\operatorname{f}^{\prime\prime}$ ranges in the interval $\left[0, 1\right]$. The first derivative of the numerator\begin{align}
\frac{\partial}{\partial x} 2 \sqrt{1 - x^2} - 2 x \arccos x = -2 \arccos x
\end{align}vanishes at $x=1$, where the numerator of $\operatorname{f}^{\prime\prime}$ assumes its minimum $0$.\qed

\balance

\printbibliography

\end{proceeding}

\end{document}